\documentclass[10pt,twocolumn,letterpaper]{article}

\usepackage{cvpr}              % To produce the CAMERA-READY version
\usepackage{graphicx}
\usepackage{amsmath}
\usepackage{amssymb}
\usepackage{booktabs}
\usepackage{multirow}
\usepackage{microtype}
\usepackage{soul}
\usepackage{algorithm}
\usepackage{algorithmic}

\usepackage[pagebackref,breaklinks,colorlinks]{hyperref}

\usepackage[capitalize]{cleveref}
\crefname{section}{Sec.}{Secs.}
\Crefname{section}{Section}{Sections}
\Crefname{table}{Table}{Tables}
\crefname{table}{Tab.}{Tabs.}

\newcommand{\buyu}[1]{{\textcolor{black}{#1}}}

\newcommand{\mc}[1]{{\textcolor{black}{#1}}}

\newlength\tikzfigwidth
\newlength\tikzfigheight

\def\txtBin{\textrm{b}}
\def\txtMc{\textrm{m}}
\def\txtReg{\textrm{c}}

\def\sa{\Theta}
\def\saBin{\sa_{\txtBin}}
\def\saMc{\sa_{\txtMc}}
\def\saReg{\sa_{\txtReg}}
\def\saBinIdx{\sa_{\txtBin{},i}}
\def\saMcIdx{\sa_{\txtMc,i}}
\def\saRegIdx{\sa_{\txtReg,i}}

\def\bevmap{x}

\def\im{I}

\def\nnFull{f}
\def\nnFeat{g}

\def\predSa{\eta}
\def\predSaBin{\predSa_{\txtBin}}
\def\predSaMc{\predSa_{\txtMc}}
\def\predSaReg{\predSa_{\txtReg}}
\def\predSaBinIdx{\predSa_{\txtBin,i}}
\def\predSaMcIdx{\predSa_{\txtMc,i}}
\def\predSaRegIdx{\predSa_{\txtReg,i}}

\def\lossSym{\mathcal{L}} % FIXME: This is generic and should be put somewhere else ...

\def\lossBCE{\textrm{BCE}}
\def\lossCE{\textrm{CE}}
\def\lossEllOne{\ell_1}

\def\realspace{\mathbb{R}} % the real space
\def\cvprPaperID{2166} % *** Enter the CVPR Paper ID here
\def\confName{CVPR}
\def\confYear{2022}

\begin{document}

\title{Weakly But Deeply Supervised Occlusion-Reasoned Parametric \mc{Road} Layouts}

\author{Buyu Liu$^{1}$ $\quad$ Bingbing Zhuang$^{1}$ $\quad$ Manmohan Chandraker$^{1,2}$ \\
$^1$NEC Laboratories America $\quad$ $^2$UC San Diego
}

\maketitle

%%%%%%%%% ABSTRACT
\begin{abstract}
We propose an end-to-end network that takes a single perspective RGB image of a complex road scene as input, to produce occlusion-reasoned layouts in perspective space as well as a parametric bird's-eye-view (BEV) space. In contrast to prior works that require dense supervision such as semantic labels in perspective view, our method only requires human annotations for parametric attributes that are cheaper and less ambiguous to obtain. To solve this challenging task, our design is comprised of modules that incorporate inductive biases to learn occlusion-reasoning, geometric transformation and semantic abstraction, where each module may be supervised by appropriately transforming the parametric annotations. We demonstrate how our design choices and proposed deep supervision help achieve meaningful representations and accurate predictions. We validate our approach on two public datasets, KITTI and NuScenes, to achieve state-of-the-art results with considerably less human supervision. 
\end{abstract}

\section{Introduction}~\label{sec:intro}
% \vspace{-0.2cm}
%\mc{MC: While the claimed contribution is both BEV and perspective semantics, the intro is written to motivate BEV.}

%Understanding road layout in top-view with perspective input is essential for real-world applications such as autonomous driving or path planning. Non-parametric representations, such as pixel-level semantics~\cite{roddick2020predicting}, generally require labor-intensive and potentially ambiguous supervision in the top-view. Having a parametric representation for top-view layout is desirable for its interpretability as it is beneficial for higher-level reasoning and for decision-making in downstream applications.

Understanding road layout from images is essential for real-world applications such as autonomous driving or path planning~\cite{sam:Gupta17a,sam:Dhiman16a,sam:Geiger14a,roddick2020predicting}, where, besides the usual perspective space outputs, top-view representations of geometry and semantics have been popular. Non-parametric representations such as pixel-level semantics \cite{roddick2020predicting} generally require labor-intensive and potentially ambiguous supervision in the top-view, for example, when dealing with occluded regions. On the other hand, parametric representations for top-view layouts are desirable for their interpretability, which is beneficial for higher-level reasoning and  decision-making in downstream applications.

\begin{figure}
 \setlength{\belowcaptionskip}{-0.35cm}
 \centering
  \includegraphics[width=1.0\linewidth]{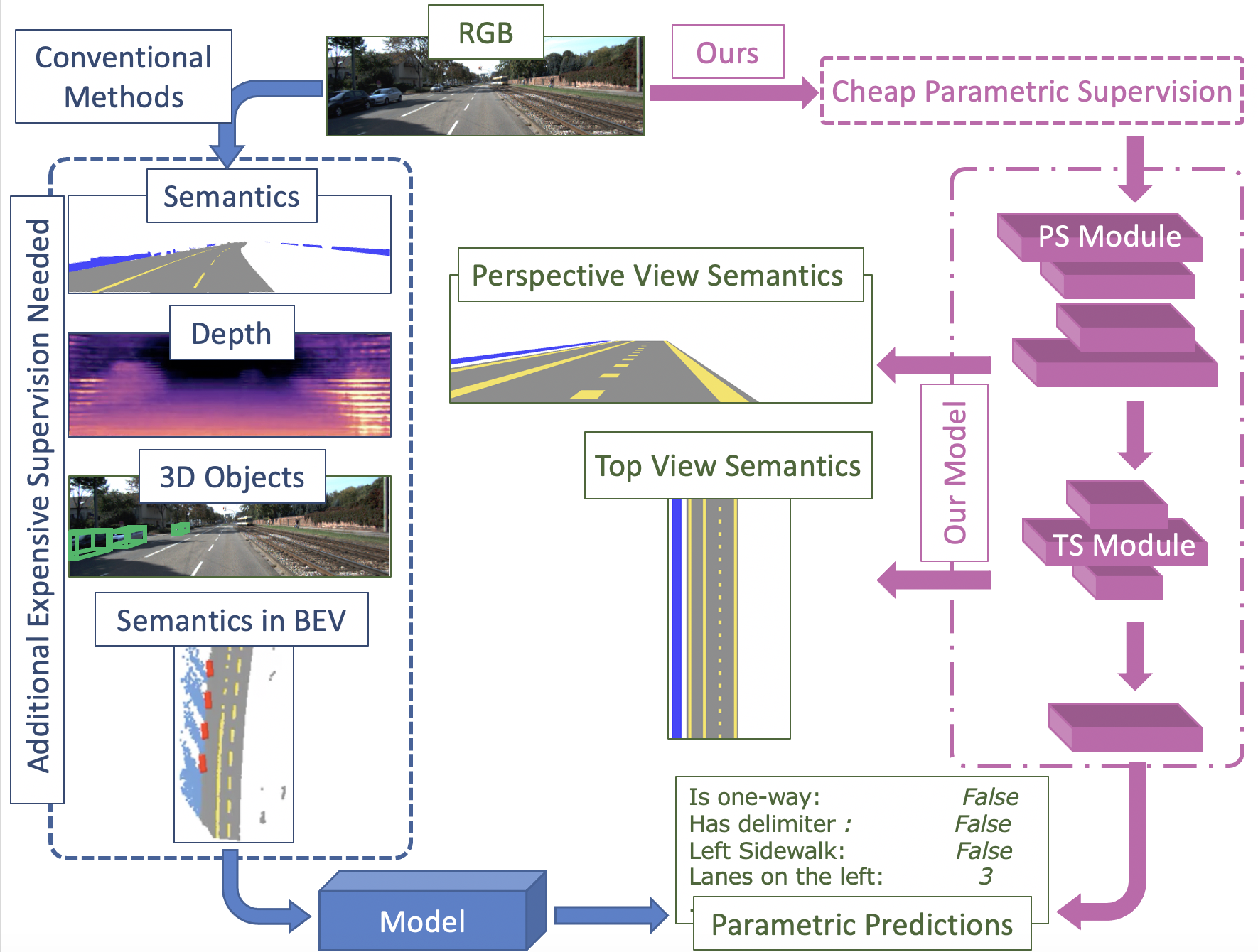}
  \caption{~\buyu{We propose an end-to-end model that inputs perspective image and outputs parametric layouts in top-view. Compared to existing methods, ours requires only the parametric layout annotations during training and achieves SOTA performance under complex road scenarios. Moreover, it generates occlusion-reasoned (see the predicted semantics on regions occluded by cars) pixel-level semantics in both perspective and top view. %Conventional method and ours are colored in blue and pink.
  } %\MC{Improve the aesthetic quality of the figure, it looks PPT-drawn and use better colors. Why is the supervision arrow from the model to the labels. Make the figure layout better -- input image and supervision on the left, model in the middle, outputs on the right (or top to bottom).} \MC{In fact, I think you should entirely re-do the teaser. The current version is just a less detailed version of Fig 2. Rather, draw a teaser that illustrates the key motivation and idea: (a) previous works need dense supervision while we only need cheap supervision, (b) inductive biases introduced through our design allows occlusion reasoning, complex geometric transformations and semantic abstractions.}  % without pixel-level human annotations.}
  %We propose an end-to-end model that requires only the parametric layout annotations during training. Our model is able to output (1) pixel-level semantics in perspective view, (2) pixel-level semantics in top-view and (3) top-view layout attribute predictions. Note that our first two representations are occlusion-reasoned (see the predicted semantics on regions occluded by cars) and are obtained without pixel-level human annotations.
  }
  \label{fig:teaser_figure}
\end{figure}

%Parametric attributes such as presence of side roads, number of lanes or distance to intersection may be easily annotated by humans given sensor inputs, 
Parametric attributes such as presence of side roads or number of lanes may be easily annotated by humans given sensor inputs, and require less effort than pixel-level semantic annotations. However, besides parametric annotations\footnote{Parametric and attribute-level annotations are used interchangeably in our paper.} in the bird's-eye-view (BEV), i.e. top-view, previous works that estimate parametric BEV layouts also require pixel-level supervision in perspective images \cite{Wang_2019_CVPR,Liu_2020_CVPR}~\buyu{or handle only very simple road layouts~\cite{sam:Seff16a}}. 
%\st{with less satisfactory performance}}. 
This paper seeks to obtain parametric BEV maps as well as pixel-level semantics in both the perspective and top views, but using only the cheaper parametric supervision on attributes.
%Compared to the former, the latter can be very time-consuming and labour-intensive.

While relying on cheap supervision only is undoubtedly a goal worth pursuing, removing the dense perspective supervision makes the problem harder. This is non-trivial, since there exists a large gap between sparse parametric supervision and dense pixel-level semantic supervision. To bridge the gap, one must reason about the underlying geometry to map the parametric supervision to top-view and get the correct semantics, even in occluded regions.

\mc{We address this challenge through two key insights. First, rather than directly regressing the parametric BEV layout from RGB image space, we introduce two intermediate steps -- a perspective semantics (PS) module and a top-view semantics (TS) module -- to predict intermediate occlusion-reasoned per-pixel perspective and BEV layouts (Fig.~\ref{fig:teaser_figure}). Second, to obtain supervision for PS/TS, a simple renderer can convert the parametric annotations to occlusion-reasoned per-pixel semantic annotations in both the BEV and perspective view, with the help of geometric transformation. This allows meaningful deep supervision \cite{pmlr-v38-lee15a,chili2017} of intermediate modules without additional annotation costs, thus weakly supervised. The weakly but deeply supervised PS and TS modules together lead to accurate parametric BEV layout by introducing inductive biases on the type of reasoning the network should perform, thereby facilitating complex tasks such as occlusion reasoning, geometric transformation and semantic abstraction that correspond to the parametric supervision.}

% We address this challenge through two key insights. First, rather than directly regressing the parametric bird-eye-view (BEV)
% %\footnote{BEV and top-view are used interchangeably in our paper.} 
% layout from RGB image space, we introduce two intermediate steps -- a perspective semantics (PS) module and a top-view semantics (TS) module -- to predict intermediate occlusion-reasoned per-pixel perspective and top-view layouts. Second, the parametric annotations can be converted to occlusion-reasoned per-pixel annotations in both the top-view and perspective view, by exploiting a simple renderer and thereafter a geometric transformation.
% %, respectively. 

% Further, our approach allows meaningful deep supervision \cite{pmlr-v38-lee15a,chili2017} of intermediate modules without additional annotation costs.
% %These annotations can then be used to deeply supervise~\cite{wang2015training} the intermediate modules.
% Our deep supervision, on the one hand, is beneficial in terms of obtaining meaningful occlusion-reasoned pixel-level semantics in both perspective and top-view without per-pixel human annotations. On the other hand, it also improves our final parametric layout predictions in BEV. Together, these two insights drive the Perspective Semantics (PS) module and Top-view Semantics (TS) module to introduce inductive biases on the type of reasoning the network should perform, thereby facilitating complex tasks such as occlusion reasoning, geometric transformation and semantic abstraction that corresponds to the parametric supervision, see Fig.~\ref{fig:teaser_figure}.

\mc{The above insights make our method simple yet highly effective, even outperforming previous methods that rely on perspective-view dense supervision for semantic segmentation. We validate our choices through state-of-the-art (SOTA) accuracies on both KITTI~\cite{sam:Geiger13a} and NuScenes~\cite{sam:NuScenes18a} datasets, achieving $47.3\%$ and $13.0\%$ F1 score.
%\MC{summarize some key numbers}. 
In extensive ablation experiments, we establish the value of the inductive biases introduced by the PS and TS modules, as well as the deep supervision through transformed parametric annotations.%\footnote{Our parametric annotations will be publicly released.}
}

% We validate the concept of the intermediate modules on KITTI~\cite{sam:Geiger13a} and NuScenes~\cite{sam:NuScenes18a}, and obtain state-of-the-art (SOTA) accuracy. We demonstrate that our method even outperforms previous methods that rely on perspective-view dense supervision for semantic segmentation.
% This verifies our intuitions on the benefits of the deeply supervised modules for the perspective and top-view semantics. 
% %Our annotations will be publicly released.

To summarize, our key contributions are:
\begin{itemize}
\item An end-to-end model for occlusion-reasoned perspective and top-view parametric layout in complex scenes.
\item Intermediate module design that incorporates inductive biases to learn occlusion-reasoning, geometric transformation and semantic abstraction.
\item Deep supervision with cheap parametric annotations~\buyu{in top view only}, rather than requiring~\buyu{additional} expensive per-pixel labeling~\buyu{in either perspective or top view}.
\item State-of-the-art results on publicly available datasets.
\end{itemize}

\begin{figure*}[t!]
 \setlength{\belowcaptionskip}{-0.35cm}
 \centering
 % \vspace{-0.5cm}
  \includegraphics[width=1.0\linewidth]{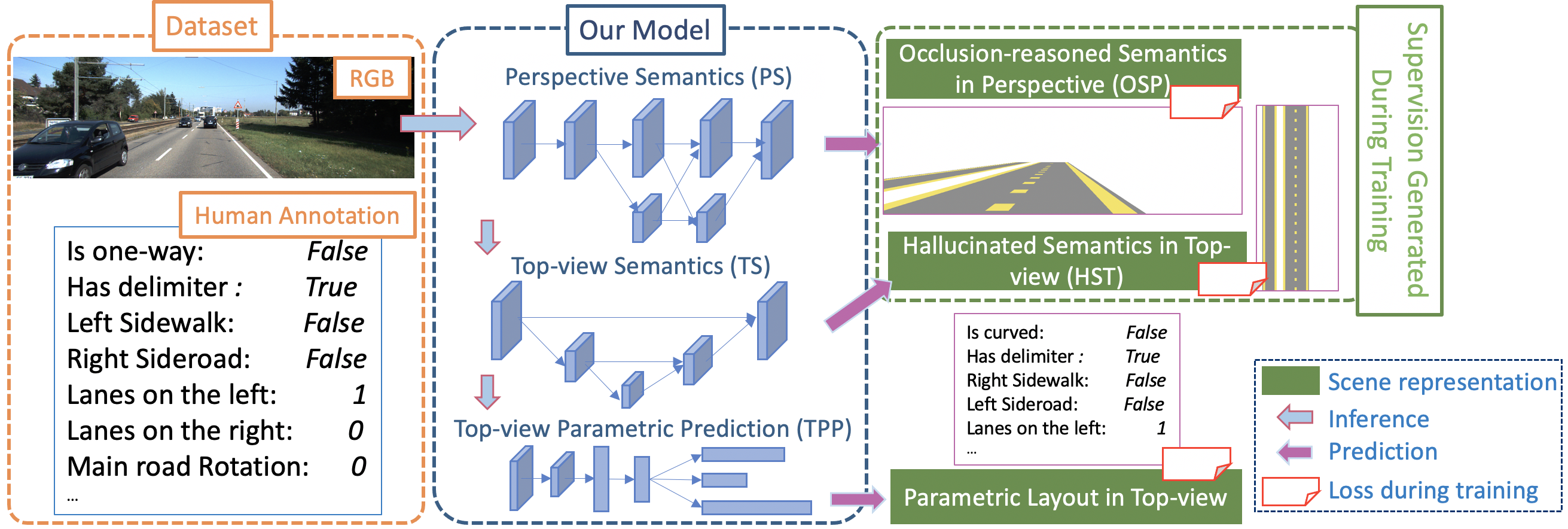}
  \caption{\textbf{Overview of our proposed framework:} Taking a single RGB as input, our model predicts (1) occlusion-reasoned semantics in perspective view, (2) hallucinated semantics in top-view and (3) parametric layout predictions in top-view, with only {\em attribute-level} annotations in top-view. This is achieved with multiple intermediate modules and deeply supervised training.}
  \label{fig:framework_overview}
\end{figure*}
\section{Related Work}~\label{sec:relawork}
3D scene understanding on outdoor scenes is an important yet challenging task.
%in computer vision. 
Applications such as robot navigation~\cite{sam:Gupta17a}, autonomous driving~\cite{sam:Geiger14a,sam:Kunze18a}, augmented reality~\cite{sam:Armeni16a} or real estate~\cite{sam:Liu15a,sam:Song18a} always require comprehensive understanding on given scenes.

%\vspace{-0.4cm}
% \noindent{\textbf{Road Scene Understanding:}}
\paragraph{Road Scene Understanding} Scene understanding for outdoor scenarios is very challenging mainly due to the lack of strong priors. To this end, non-parametric approaches have been proposed~\cite{sam:Guo12a,Tighe_2014_CVPR,sam:Tulsiani18a}, where layered representations~\cite{zhan2020self,dhamo2019object} are utilized to reason about the geometry as well as semantics in occluded areas. Other typical non-parametric representations in perspective view are joint pixel-level semantics and depth~\cite{liu2010single}, pixel-level semantics and geometric labels~\cite{gould2009decomposing}. In contrast, parametric approaches provide abstract understanding, such as road scene attributes~\cite{sam:Seff16a,sam:Geiger14a} and graph-based representation~\cite{sam:Kunze18a}. Perhaps~\cite{Wang_2019_CVPR,Liu_2020_CVPR} are the most recent works that are able to handle complex road layout, e.g. multiple lanes and different types of intersections. Our work follows the parametric representation proposed in these methods. Unlike~\cite{Wang_2019_CVPR,Liu_2020_CVPR} that request additional information, e.g. models~\cite{he2016deep} pre-trained with dataset-specific per-pixel semantics, depth and 3D objects~\cite{sam:Geiger13a}, to map semantics to top-view as pre-processing, our model is end-to-end trainable that directly takes RGB as input. More importantly, we exploit deep supervision~\cite{pmlr-v38-lee15a,chili2017} by introducing meaningful intermediate modules (PS and TS), with which we are able to obtain occlusion-reasoned pixel-level semantics in both perspective and top-view without per-pixel human annotations. It is also beneficial in terms of improving final parametric layout predictions. Though focusing on single image for now, 
%Please also note that although we focus only on single images, 
our model can be easily extend to video-version by introducing spatio-temporal graphical model~\cite{liu2015multiclass,Wang_2019_CVPR},  LSTM~\cite{feichtenhofer2017spatiotemporal,simonyan2014two} or FTM~\cite{zhu2017flow,zhu2017deep,vu2018memory}
\paragraph{Scene Understanding in Top-view}
Top-view representations~\cite{NEURIPS2019_ba2f0015,Lu2019icra-ral,pan2019crossview,mani2020autolay} can be more beneficial when occlusion relationships are desired, e.g., two objects cannot occupy the same position in top-view while they can potentially occlude each other in perspective view. Such intuition is widely exploited in 3D object localization literature~\cite{wang2019pseudo} where camera to top-view projection is fulfilled with the help of depth estimation and 2D detection in perspective view. Although~\cite{roddick2018orthographic} proposes an end-to-end trainable model that explicitly exploits the perspective to top-view projection to perform 3D localization task, the performance of this method is not comparable to~\cite{wang2019pseudo} due to the lack of explicit depth-aware re-projection. As for general scene understanding, the initial steps are taken in~\cite{sam:Schulter18a,sam:Sengupta12a}. However, due to the lack of ground truth, no quantitative evaluation is performed in~\cite{sam:Schulter18a}. More recent work~\cite{roddick2020predicting,philion2020lift} extends~\cite{roddick2018orthographic} and predicts top-view semantic map from single monocular image or multiple streams of images.~\buyu{A graph like parametric representation is introduced in~\cite{Can2021ICCV} for road layout estimation as well as oriented bounding boxes for road participants. However, such representation misses important semantics such as crosswalk, sidewalk and lane directions. And it further requires HD-map, GPS and human annotations to train such a model.}%Jonah and Sanja~\cite{philion2020lift} later on propose a more general model to improve the representation power of~\cite{roddick2020predicting}. 
%Note that scene representations in
~\buyu{In contrast to non-parametric approaches~\cite{roddick2020predicting,philion2020lift,mani2020monolayout,Yang_2021_CVPR,pan2019crossview} that require expensive per-pixel supervision in top or perspective view} and focus on predicting semantics on visible regions, our method aims to predict parametric layouts in BEV and also provides occlusion-aware non-parametric representations in both BEV and perspective view as by-products. All these meaningful representations are obtained without per-pixel human annotations but relying on cheap parametric annotations.

\section{Our Framework}~\label{sec:model}
Our model consists of three modules. (1) The perspective semantics (PS) module inputs the RGB image and outputs the occlusion-reasoned pixel-level semantics in perspective view (OSP). (2) The top-view semantics (TS) module projects OSP into top-view and learns to hallucinate or complete pixel-level top-view semantics on out-of-view as well as noisy regions, which we refer to as hallucinated semantics in top-view (HST). %\BB{What does H,S,T mean respetively? Hallucinate semantic top-view?} 
%in this paper. 
(3) The top-view parametric prediction (TPP) module takes the HST and predicts road layout related attributes in top-view. Fig.~\ref{fig:framework_overview} gives an overview of the proposed method.~\buyu{Network architectures are borrowed from \cite{wang2020deep,ronneberger2015u,Wang_2019_CVPR} \mc{and described in Sec.~\ref{sec:exp} and supplementary. We focus in this section on describing our main contributions that allow}}
% Note that we borrow existing network architectures~\cite{wang2020deep,ronneberger2015u,Wang_2019_CVPR}, as our main contributions instead lie in 
effectively exploiting weak supervision with cheap parametric-level human annotations.
We detail each module in Sec.~\ref{sec:model_model}, the training process and the generation of intermediate pixel-level semantic annotations in Sec.~\ref{sec:model_train}. 

\subsection{Full Model}~\label{sec:model_model}
Consider a dataset $\mathcal{D} = \{\im, \sa\}_{i=1}^{N}$ of $N$ samples, where $\im \in \realspace^{~\textit{H} \times \textit{W} \times 3}$ are RGB perspective images and $\sa$ denote the corresponding scene attributes obtained from human annotations. We further generate $x^p, \bevmap$ automatically for each sample where $x^p \in \realspace^{~\textit{H} \times \textit{W} \times (C+1)}$ denotes semantic segmentation map in perspective view and $\bevmap \in \realspace^{\textit{h}\times~\textit{w}\times(C+1)}$ denotes top-view semantics. $C=4$ denotes the number of % \st{background} 
layout categories (``road", ``sidewalk", ``lane boundaries", ``crosswalks") and we also include one foreground class. % \MC{Better to call them layout categories instead of background?}
%$\textit{h}$ and $\textit{w}$ are the spatial dimensions. 
We refer the readers to Sec.~\ref{sec:model_train} for more details about the data generation process.
%Let us first assume that we have collected a data set $\mathcal{D} = \{\im, x^p, \bevmap, \sa\}_{i=1}^{N}$ of $N$ samples. Specifically, $\im \in \realspace^{~\textit{H} \times \textit{W} \times 3}$ is the RGB perspective image, with height $\textit{H}$ and width $\textit{W}$. $x^p \in \realspace^{~\textit{H} \times \textit{W} \times (C+1)}$ denotes semantic segmentation map in perspective view, with $(C+1)$ semantic categories (``road", ``sidewalk", ``lane boundaries", ``crosswalks" and ``foreground"). $C=4$ denotes the number of background categories.~\buyu{In practice, $\sa$ is obtained with human annotations while $x^p, \bevmap$ are generated ourselves. We refer the readers to Sec.~\ref{sec:model_train} for more details.} We further denote semantic top-views as $\bevmap \in \realspace^{\textit{h}\times~\textit{w}\times(C+1)}$, with spatial dimensions $\textit{h}\times\textit{w}$, containing the same ${C+1}$ semantic categories. Finally, we denote scene attributes as $\sa$ for each data sample.
% for each data sample, we denote its corresponding scene attributes as $\sa$. 
Our full model is defined as:
\begin{equation}
    \sa = \nnFull^{\textrm{full}}(\im)=(\nnFull^{tpp} \circ \nnFull^{\textrm{ts}} \circ \nnFull^{\textrm{ps}})(\im) \;,
\end{equation}
where $\circ$ defines a function composition. $\nnFull^{\textrm{ps}}$, $\nnFull^{\textrm{ts}}$ and $\nnFull^{tpp}$ %\BB{put tpp on f too} 
correspond to our three modules PS, TS, and TPP.

\begin{figure}[t]
 \setlength{\belowcaptionskip}{-0.35cm}
 \centering
 \includegraphics[width=1.0\linewidth]{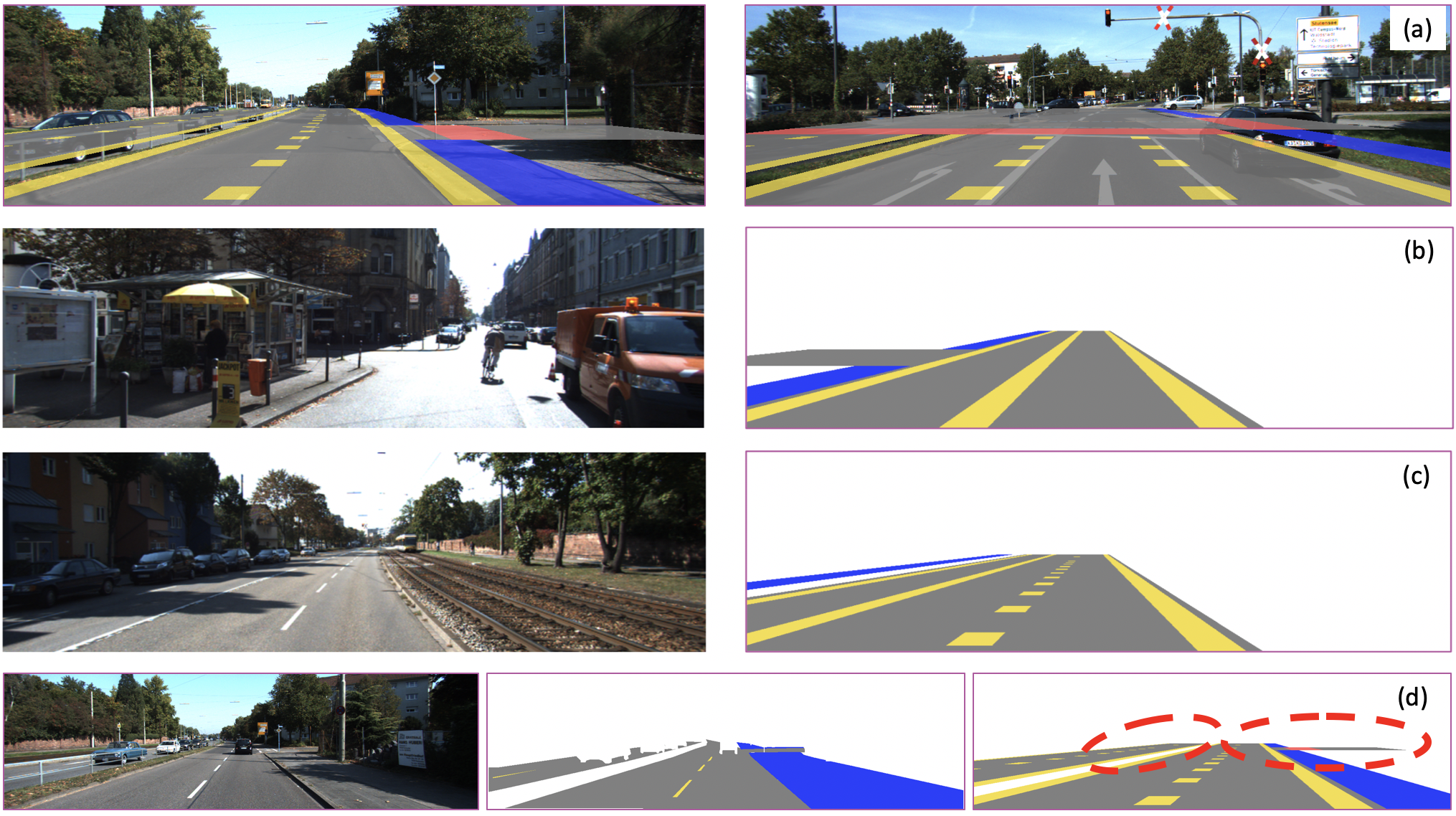}
  %\includegraphics[trim=0 0 0 200,clip,width=1.0\linewidth]{figs/persp_gt.png}
  %\includegraphics[width=1.0\linewidth]{figs/gt_samples.png}
  % \vspace{-0.4cm}
  \caption{%\BB{The way to overlay two images in the first row makes both images unclear. You can instead use alpha blending} 
  We overlay output and input of PS module in (a). Examples of inputs (left) and target outputs (right) of PS module are provided in (b) and (c). PS module aims to predict both visible and occluded background classes. (d) demonstrates the input image, semantics of visible regions and our target output from left to right. And we highlight the occluded regions in red. %\BB{put (a)(b)(c)(d) in the left of figure, and refer them as Fig.3(a) etc, not the first row of Fig.3 etc.}%\BB{third row is not referred anywhere in the paper}
  %As can be seen in this figure, although not $100\%$ accurate, we can obtain satisfactory occlusion-reasoned semantics in perspective.
  }
  \label{fig:gt_persp}
\end{figure}

\vspace{-0.2cm}
\paragraph{Perspective semantics module}
The PS module predicts per-pixel occlusion-reasoned semantics in the perspective view (OSP). Unlike traditional semantic segmentation models (e.g. ~\cite{wang2020deep,chen2014semantic}) %\BB{why citing [33] rather than more well-known semantic model like DeepLab?} 
that predict semantics on visible pixels only, our module focuses on predicting both visible and occluded layout classes (See Fig.~\ref{fig:gt_persp}(d)). Such occlusion reasoning is also demonstrated in Fig.~\ref{fig:gt_persp}(b) and (c). As shown, we aim to predict road semantics in the top-view despite that they are occluded, e.g. by cars or buildings, in the perspective view.  %\BB{foreground is ambiguious here, it usually refer to cars, pederstrains etc, not road. foreground --- road}
%Specifically, this module aims to predict where the foreground classes really lie in the perspective view, even if they have been occlude by background class. 

%Taking the last row in Fig.~\ref{fig:gt_persp} %, where several cars are parking along side the road, 
%as an example.
%to demonstrate why our predictions are occlusion-reasoned.
%As human beings we are aware that cars are parking on the~\textit{Road} and based on the contextual information, they also occlude partial~\textit{Sidewalk}. Given the fact that our ultimate goal is to predict the road layout, instead of predicting the semantics of visible regions, we are more interested in the location road layout related semantics, or the 4 background classes. 
%In this example,
%Instead of predicting~\textit{Car}, more desired predictions for these "car" regions would be~\textit{Sidewalk} \BB{is Sidewalk occluded by car?} and~\textit{Road} in the top-view road layout prediction task. An even harder example is shown in the second row of Fig.~\ref{fig:gt_persp}, where road area is occluded by buildings in perspective view. 
%Such scenario happens frequently when ego car approaches the intersections. 
%and can be critical for us to understand the road layout in top-view. 
Compared to conventional semantic segmentation problem, ours is more challenging in terms of both data and model training. As for data, the semantic ground-truth on occluded regions can be ambiguous, hence difficult and time-consuming to annotate accurately in pixel-level. %\mc{MC: There was the issue in CVPR reviews around LIDAR for ground truth distance annotations. How is it being addressed now?} 
For instance, it takes more than 20 minutes to annotate only the visible regions on KITTI images while in comparison, parametric annotation in BEV takes about 20s for an image~\cite{Wang_2019_CVPRworkshop}. We refer the readers to Sec.~\ref{sec:exp} and supplementary for annotation details. %\BB{20s seems unrealistically fast for distance measurement, how Lidar is used here?} 
For model training, the PS module predicts semantics in invisible/occluded regions, which again requires dealing with ambiguity. For instance, regions occluded by a foreground instance, e.g. building, can be either another building or road. This requires the module to learn to predict semantics with contextual cues rather than fully relying on local visible information.  

%\BB{This paragraph is good at explaining occlusion, but too verborse and should be more concise, e.g. removing "parking" etc.}~\buyu{Will make them more concise after BB added more details on projection and back-projection.}
% we ask the network to predict the semantics that are potentially occluded by various road participants or background, such as pedestrians, cars and buildings. 
%An even harder case would be we would like to recover a sideroad that has been occluded by foreground, such as a building, in perspective.

%\vspace{-0.2cm}

%Assuming that such desired ground-truths, $x^p$, are available in our dataset, %we then follow the structure  %of~\cite{wang2020deep}, or 
%of HRNetV2-W18~\cite{wang2020deep} as our semantic segmentation backbone as it achieves very good trade-offs between accuracy and efficiency. 
Formally, given an image $\im$, %\st{$\im \in \realspace^{~\textit{H}\times\textit{W}\times3}$}, 
the PS module outputs $x^p$ 
% \st{$\in \realspace^{~\textit{H}\times\textit{W}\times(C+1)}$, which denotes} 
encoding the probability of each pixel belonging to a specific category: 
%We consider $C=4$ road layout related semantic categories as also %, where $H$ and $W$ are the height and width of perspective images and $C$ equals to 4 and represents the number of layout related classes, or~\textbf{Road,Land boundary, Sidewalk} and~\textbf{Crosswalk}. 
%And we also 
%take the foreground class into consideration. 
%Given the input and output in perspective view, w
% \st{We define the PS module as:}
%
\begin{equation}
    x^p = \nnFull^{\textrm{ps}}(\im) \;.
\end{equation}
% where $\nnFull^{\textrm{ps}}$ denotes the PS module. %See the figure below for more details of the structure of this module.~\buyu{Figure that demonstrate the structure of module 1 is needed here.}
% \vspace{-0.5cm}
% \noindent{\textbf{Top-view Semantics Module}}
\paragraph{Top-view semantics module}
Our second module, i.e. the top-view semantics module, takes as input the OSP and learns to explicitly project the semantics in perspective view to top-view. Given camera intrinsics, the projection could be done if the depth estimation is available, say, via a depth network. However, standard single image depth networks (e.g.~\cite{liu2010single,godard2017unsupervised,garg2016unsupervised}) typically do not reason about depth in occluded regions, which is nevertherless required for our occlusion-aware projection. In addition, resolution is low for distant regions and thus may lead to sparser/noisier semantics in top-view. Lastly, top-view semantics on close-by regions can be incomplete due to limited field of view. Instead, we propose a two-step projection through an initial geometric transformation $\nnFull^{\textrm{trans}}$ and a learned hallucination module $\nnFull^{\textrm{halln}}$:
\begin{equation}
    \bevmap = \nnFull^{\textrm{ts}}(x^p)=(\nnFull^{\textrm{halln}} \circ \nnFull^{\textrm{trans}})(x^p).
\end{equation}

\noindent{\textit{Transformation module}.} In view of these issues, we first make use of the prior that the road forms nearly a plane, which facilitates an initial projection without requiring depth estimation. 
%\st{We term this step a transformation module.}
%top-view. %\BB{Buyu, please change this second issue accordingly}. 
%we propose to first project the semantics in perspective view to top-view with plane assumption and then learn to hallucinate the projection results. 
%Specifically, our TS module consists of a transformation module and a hallucination module. 
We assume known camera intrinsics and extrinsics w.r.t. the ground plane; this is a mild assumption since they could be obtained via calibration~\cite{hartley2003multiple} in advance. As such, it is well-known that one can back-project each pixel in the perspective view to the BEV view and vice versa~\cite{hartley2003multiple}. 

% \st{Second, we introduce a hallucination module to address the sparse/noisy semantics and incomplete predictions on far-away and close-by regions in top-view.}

\begin{figure}[t]
 \setlength{\belowcaptionskip}{-0.35cm}
 \centering
  \includegraphics[width=1.0\linewidth]{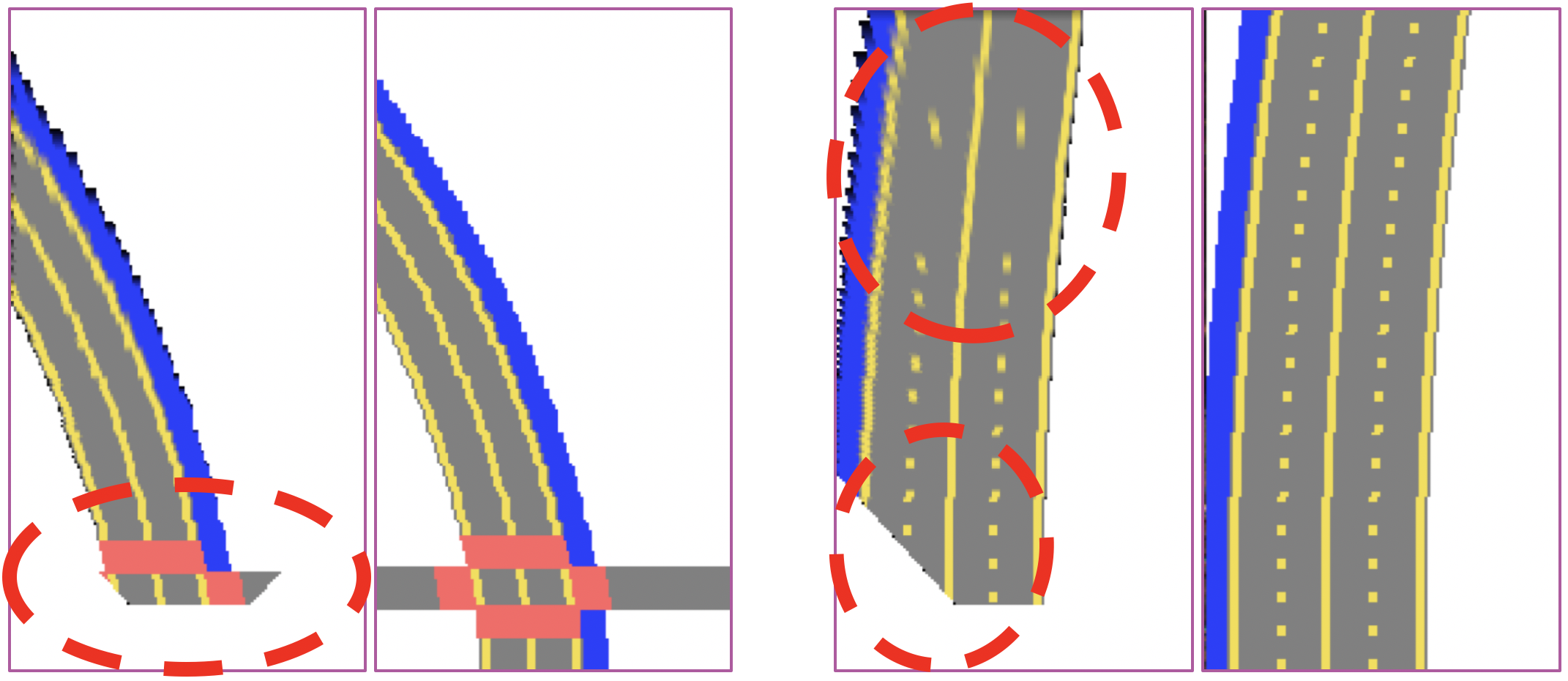}
  % \vspace{-0.4cm}
  \caption{Two sets of examples for input and output of hallucination module. Our module aims to recover the far away sparse semantics (right) and hallucinate close-by areas with limited view (left).%\BB{Highlight the regions where we want readers to look, with red circles.}
  }
  \label{fig:bev_hallucination}
\end{figure}

\noindent{\textit{Hallucination module}.} After the transformation module that maps the OSP to top-view, the hallucination module then learns to predict the unseen far away regions as well as recover the noisy semantics with contextual information in top-view. 
%We borrow the structure from~\cite{ronneberger2015u} and utilize a shallow version, e.g. 5-layer encoder and decoder, as our hallucination module. 
Note that our input and output of hallucination module are both of the size $\textit{h}\times\textit{w}\times(C+1)$. Fig.~\ref{fig:bev_hallucination} visualizes two sets of inputs and outputs of this module. Compared to inputs generated with ground-truth OSP, the target HST improves at far away (right) regions as well as close-by areas where predictions are sparse (left).   
%We refer the readers to the last column of Fig.~\ref{fig:bev_hallucination} as examples of targeted hallucinated results.
%\BB{Can we have an example comparing before and after hallucination side by side?}

% In summary, we define the TS module as:
% \begin{equation}
%     \bevmap = \nnFull^{\textrm{ts}}(x^p)=(\nnFull^{\textrm{halln}} \circ \nnFull^{\textrm{trans}})(x^p) \;,
% \end{equation}
% where %$\circ$ defines a function composition. 
% $\nnFull^{\textrm{halln}}$ and $\nnFull^{\textrm{trans}}$ are the hallucination and transformation module, respectively.

% In short, the encoder consists of the repeated application of two 3x3 unpadded convolutions, each followed by a rectified linear unit (ReLU) and a 2x2 max pooling operation with stride 2 for downsampling. At each downsampling step we double the number of feature channels.  The deconder upsamples the feature map with a 2x2 convolution (“up-convolution”) that halves the number of feature channels, a concatenation with the correspondingly cropped feature  map  from  the  contracting  path,  and  two  3x3  convolutions,  each  fol-lowed by a ReLU. The cropping is necessary due to the loss of border pixels in every convolution. At the final layer a 1x1 convolution is used to map each 64-component feature vector to the desired number of classes. In total the networkh as 23 convolutional layers.
% \vspace{-0.4cm}
%\noindent{\textbf{Top-view Parametric Prediction Module}}
\paragraph{Top-view parametric prediction module}
Given the hallucinated semantics in top-view (HST), our next step is to predict the layout attributes through the top-view parametric prediction (TPP) module that maps the HST $\bevmap$ into the scene model parameters $\sa$. As aforementioned, we follow the attribute definitions in~\cite{Wang_2019_CVPR,Liu_2020_CVPR}. Our $\sa$ consists of three groups: $\saBin$ for 14 binary, $\saMc$ for 2 multi-class and $\saReg$ for 10 continuous attributes of the scene model, respectively. Binary attributes consist of information such as whether the road is one-way or not. Number of lanes on the left hand-side of the ego car is an example of multi-class attributes and distance to right side-road can be one of the continuous attributes. More details can be found in supplementary materials. 
%We refer readers to supplementary materials for more details on scene attribute definition.
%For each feasible combination of these attributes, we will be able to reconstruct the road layout in top-view. As shown in~\cite{Wang_2019_CVPR,Liu_2020_CVPR}, this parametric representation is able to handle diverse complex road layout in top-view.
%Given the input and output, we define our mapping as:
%
Our TPP module is defined as:
\begin{equation}
    \sa = \nnFull^{tpp}(\bevmap) = (\nnFull \circ \nnFeat)(\bevmap) \;,
\end{equation}
%
%\BB{h is already used in hxw, so use another symbol.}
where $\nnFull$ and $\nnFeat$ are respectively multi-layer perceptron (MLP) and convolutional neural networks. %, with weights $\nnAttrWgts$ and $\nnFeatWgts$ respectively, that we want to learn. 
%$\nnFeat$ is a convolutional neural network (CNN) that converts top-view semantics $\bevmap \in \realspace^{\textit{h}\times~\textit{w}\times(C+1)}$ into a 1-dimensional feature vector $\bevmapFeat \in \realspace^{\bevmapFeatDim}$.
%Later, a multi-layer perceptron (MLP) $\nnAttr$ is introduced to predict the scene attributes $\sa$ given $\bevmapFeat$.
Note that similar to~\cite{Wang_2019_CVPR}, this module is also able to exploit rich simulated data during training, but we leave this extension to future work.

% Our objective is that $\nnAttr \circ \nnFeat$ works well on real data, while we want to leverage the rich and large set of annotations from simulated data during training.  The intuition behind our design is to have a unified $\nnFeat$ that maps semantic top-views $\bevmap$ of different domains into a common feature representation, usable by a domain-agnostic classifier $\nnAttr$.  To realize this intuition, we define supervised loss functions on both real and simulated data and leverage domain adaptation techniques to minimize the domain gap between the output of $\nnFeat$ given top-views from different domains.

\subsection{Model Training}~\label{sec:model_train}
% That's where the overview figure was (I think)
%We demonstrate our full model in the precceding section with the assumption that supervisions are all available for each module. 
%\BB{For demonstrating TS and PS, why not use same example input and demonstrate sequentially the TS and PS? This way will make it flow smoothly and clearly. Instead, now it is on different RGB, and split into two figures, Fig.5 and Fig.3.}
Following our above description of intermediate modules assuming supervision available,  we describe in this section the generation of such supervision with only annotations for parametric layout $\sa$, as well as deep supervision training. Instead of training the full model in an end-to-end manner from scratch, we adopt a multi-stage training protocol. We first pretrain all three modules, and then jointly train the full model in an end-to-end manner. Empirically, end-to-end training provides a $1\%$ performance improvement in our experiments. Our full loss function $\lossSym$ is defined as:
\begin{equation}
\begin{split}
    \lossSym= \lambda \lossSym^{\textrm{tpp}} + \gamma \lossSym ^{\textrm{ts}} + \beta  \lossSym^{\textrm{ps}},
\end{split}
\end{equation}
where $\lambda$, $\gamma$ and $\beta$ are the weights for each module.

%\subsection{top-view Parametric Prediction Module}
%Since $\sa$ and $\im$ are now available~\cite{Wang_2019_CVPR}.  We can introduce the 
%Denoting the $\dsReal$ as the real data, $\sa = \{ \saBin, \saMc, \saReg\}$ is the ground truth of all samples in this dataset.

%\noindent{\textbf{Loss Functions on Top-view Parametric Prediction Module:}}

%\vspace{0.2cm}
%\noindent{\textbf{Top-view Parametric Prediction Module}} 
\paragraph{Top-view parametric prediction module} Since $\sa$ and $\im$ are already available, we define the loss function of TPP as: %in~\cite{Wang_2019_CVPR}, we define the loss function of parametric prediction as:% a supervised loss as
\begin{equation}
    \lossSym^{\textrm{tpp}}= \sum_{i=1}^{N} \lossBCE(\saBinIdx, \predSaBinIdx) + \lossCE(\saMcIdx, \predSaMcIdx) + \lossEllOne(\saRegIdx, \predSaRegIdx) \;,
\end{equation}
%
%\BB{Why is there a 'r' everywhere and not explained? what is $l_1$?}
where (B)CE is the (binary) cross-entropy loss and $\lossEllOne$ denotes L1 loss.  $\{\sa,\predSa\}_{\cdot,i}$ denotes the $i$-th sample in the data set. For regression, we discretize continuous variables into 100 bins by convolving a dirac delta function centered at $\saReg$ with a Gaussian of fixed variance. 
%Such multi-modal predictions allow potential application of our model output with subsequent probabilistic graphical models. 
%In addition, we may also exploit simulate data in this module as proposed in~\cite{Wang_2019_CVPR} and we choose to leave this trivial extension to future work. 
% \vspace{-0.3cm}

\begin{figure}[t]
 \setlength{\belowcaptionskip}{-0.35cm}
 \centering
  \includegraphics[width=1.0\linewidth]{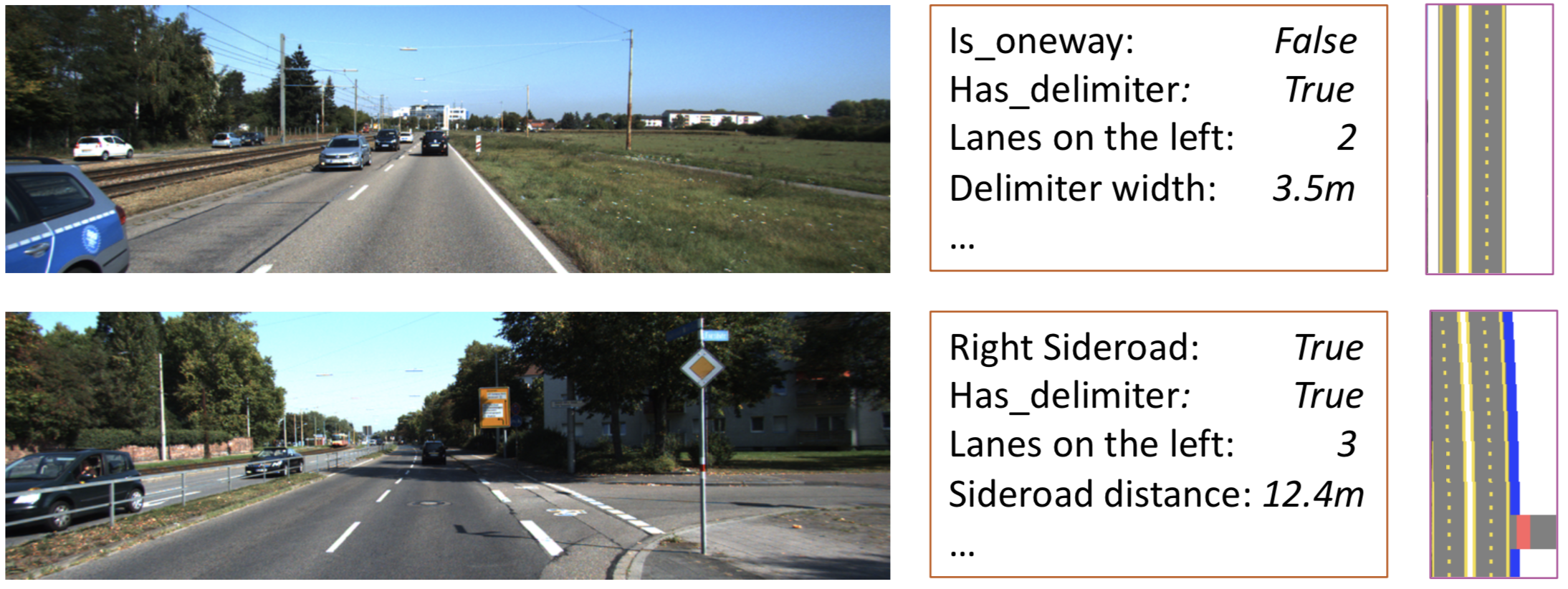}
  \caption{Examples of rendered ground-truth for TS module. From left to right: RGB, parametric human annotations and rendered pixel-level semantics in top-view. 
  }
  \label{fig:bev_rendering}
\end{figure}

%\vspace{0.2cm}
%\noindent{\textbf{Top-view Semantics Module}}
\paragraph{Top-view semantics module}
Unlike the straightforward design in parametric space, our TS module requires per-pixel supervision in top-view. To this end, we propose to exploit a rendering function that generates pixel-wise semantics from parametric annotations. Specifically, for each $\sa$, we render a map $\bevmap$. Some examples of our paired $\{\bevmap,\sa\}$ are in Fig.~\ref{fig:bev_rendering}, which shows that our rendered $x^p$ accurately reflects the layout of the road in top-view. Since we only need on parametric abstractions, our renderer can be implemented using simple Python code, rather than the complex machinery of physics-based image renderers. We refer the readers to supplementary materials for more details on our renderer and the generation process.
%\noindent{\textbf{Loss Functions on Top-view Semantics Module}}
The loss function for TS module is defined as:
\begin{equation}
    \lossSym^{\textrm{ts}}= \sum_{i=1}^{N} \lossCE(\bevmap_i, \hat{\bevmap_i})
\end{equation}
where $\hat{\bevmap_i}$ and $\bevmap_i$ denotes the predictions and the rendered ground-truth of the top-view semantics of $i$-th sample in $\mathcal{D}$.
%\vspace{-0.3cm}

%\vspace{0.2cm}
%\noindent{\textbf{Perspective Semantics Module}}
\paragraph{Perspective semantics module}
Obtaining the top-view semantics $\bevmap$, we can project~\cite{hartley2003multiple} it to perspective view with camera parameters as well as plane assumption. 
%Again, this is achieved by computing the homography~\cite{hartley2003multiple} between the perspective and BEV view. 
%Here are some examples of input image, perspective semantics as well as top-view semantics. As can be seen in this figure, our back projection does a great job and can indeed recover the occluded regions in perspective view. ~\buyu{Example figures of rendering function to be demonstrated here.}
%We refer the readers to the first row of Fig.~\ref{fig:gt_persp} 
%as examples of the generated ground-truth for perspective module. 
We demonstrate the effectiveness of our projection in Fig.~\ref{fig:gt_persp}(a).
%Fig.~\ref{fig:gt_persp}(a) demonstrates the effectiveness of our mapping. Our back projection can indeed recover the occluded regions and aligns the semantics well with respect to input image in perspective view.
%\noindent{\textbf{Loss Functions on Perspective Semantics Module}}
Similarly, the loss function for the PS module is defined as:
\begin{equation}
    \lossSym^{\textrm{ps}}= \sum_{i=1}^{N} \lossCE(x^p_i, \hat{x}^p_i)
\end{equation}
where $\hat{x}^p_i$ and $x^p_i$ denotes our predictions and the back-projected ground-truth of the perspective semantics of sample $i$ in $\mathcal{D}$.
  
\section{Experiments}~\label{sec:exp}

\noindent \textbf{Datasets and model details~~~} We validate our ideas on KITTI~\cite{sam:Geiger13a} and NuScenes~\cite{sam:NuScenes18a}, utilizing the annotation and data split in~\cite{Liu_2020_CVPR}. Please refer to \cite{Liu_2020_CVPR} and our supplementary for details in road layout attributes annotation. 
% ~\footnote{It includes around 17000 annotations for KITTI and about 1200 annotations for NuScenes in terms of scene layout annotation.}. 
$h$ and $w$ are set to 256 and 128, presenting a 60m $\times$ 30m space in real world. Camera parameters are available in the original datasets  %\cite{sam:Geiger13a,sam:NuScenes18a} 
through calibration. Weights $(\lambda,\gamma,\beta)$ are set experimentally on validation set.~\buyu{As for $f^{ps}$, we use HRNetV2-W18~\cite{wang2020deep} as the backbone as it achieves very good trade-offs between accuracy and efficiency. As for $f^{halln}$, we utilize a shallower version of~\cite{ronneberger2015u}, e.g. 5-layer encoder and decoder. Finally, $\nnFull$ is implemented as a multi-task network with three separate predictions $\predSaBin$, $\predSaMc$ and $\predSaReg$ for each of the parameter groups $\saBin$, $\saMc$ and $\saReg$ of the scene model. And $\nnFeat$ is introduced for feature extraction. Note that our method does not depend on the specific details of these sub-modules but is generally applicable if this three-stage architecture holds.}

%\buyu{Note that unlike previous works \cite{Wang_2019_CVPR,Liu_2020_CVPR} that requires thousands of images to be annotated to train segmentation networks, LiDAR to train dense depth prediction network and 3D object annotations to obtain contextual cues, our model requires only cheap human annotations in parametric space.}
\begin{table*}[t!]
  \setlength{\belowcaptionskip}{-0.35cm}
  \centering\small
  \begin{tabular}{l|c|c|c|c|c|cccc}
  \hline
              &
              \multicolumn{5}{c|}{Supervision Required} &\multicolumn{4}{c}{KITTI~\cite{sam:Geiger13a}} \\
  \hline
  Method      & Parametric & Depth & Semantics & Simulated & Video+Object &  Accu.-Bi. $\uparrow$ & Accu.-Mc. $\uparrow$ & MSE $\downarrow$ & F1 $\uparrow$ \\
  \hline
  RGB~\cite{sam:Seff16a,he2016deep}    & \checkmark & & & & & .811 & .778 & .230& .176\\
  \hline
  RGB~\cite{sam:Seff16a,he2016deep}+D   & \checkmark & \checkmark & & & & .818  & .819  & .154&  .109\\
  BEV~\cite{sam:Schulter18a}  & \checkmark & \checkmark & \checkmark & & & .820& .797& .141& .324 \\
  H-BEV+DA~\cite{Wang_2019_CVPR} & \checkmark & \checkmark & \checkmark & \checkmark & & \textbf{.834} & .831 & \textbf{.134} & .435\\
  BEV-J-O~\cite{Liu_2020_CVPR} & \checkmark & \checkmark & \checkmark & & \checkmark & .831 & \textbf{.837}  & .142 & \textbf{.494} \\  
  \hline
  Ours & \checkmark & & & & & \underline{.833} & \underline{.832} & \underline{.140} & \underline{.473} \\
  \hline
\end{tabular}
  %\vspace{-0.2cm}
  \caption{Performances on single image based road layout prediction on KITTI. we observe that our method outperforms $RGB$ when having the same model setting. In addition, our results are comparable to other SOTA ($H$-$BEV$+$DA$ and $BEV$-$J$-$O$) but with far less human annotations required.}
  \label{tbl:kitti_single_img}
\end{table*}
% \vspace{-0.5cm}
\begin{table}[t]
  \setlength{\belowcaptionskip}{-0.35cm}
  \setlength{\tabcolsep}{4pt}
  \centering\small
  \begin{tabular}{l|c|c|c|c}
  %\hline
  %            & \multicolumn{3}{c}{Annotation Time (s)} \\
  \hline
  Time     & Binary &  Multiclass & Continuous & Total\\
  \hline
  Random images & 24.3 & 5.1 & 25.7 & 55.1\\
  \hline
  Video frames & \multicolumn{4}{c}{20.2}\\   
  \hline
\end{tabular}
  \caption{Average annotation time (sec.) on KITTI dataset.}
  \label{tbl:annotation_table}
\end{table}

%\vspace{-0.2cm}
%\noindent{\textbf{Parametric annotations}}
\paragraph{Cost for parametric annotations}
We summarize the annotation time for each type of supervision in Tab.~\ref{tbl:annotation_table}. Unlike non-parametric annotations such as pixel-level semantics that require several dozens of minutes per frame, our parametric annotations require less than a minute per frame. Moreover, this time is heavily amortized across a video sequence to just around 20 seconds on the KITTI dataset, since parametric attributes change predictably across consecutive frames. Binary and multiclass attributes (such as presence of side-road, or number of lanes) change less frequently and their annotations can often be inherited from previous frames. Further, continuous attributes (such as distance to intersection) typically change smoothly across frames, which facilitates annotation. We refer the readers to supplementary materials for more details. 
%Our annotations will be publicly released.

%.simple parametric annotations. 
%\BB{point out that our annotation needs Lidar.}
% for each module.
%We validate our ideas on two data sets, KITTI~\cite{sam:Geiger13a} and NuScenes~\cite{sam:NuScenes18a}. Similarly, we utilize the annotated data in~\cite{Liu_2020_CVPR}, which includes around 17000 annotations for KITTI~\cite{sam:Geiger13a} and about 1200 annotations for NuScenes~\cite{sam:NuScenes18a} in terms of scene layout annotation. Note that unlike previous work~\cite{Wang_2019_CVPR,Liu_2020_CVPR} that requires thousands of images (1243 images for training and 382 for testing on KITTI) to be annotated to train off-the-shelf segmentation network and also LiDAR images to train dense depth prediction network, our model requires only simple parametric annotations.
%\vspace{-0.2cm}
% \noindent{\textbf{Evaluation metrics}}
\paragraph{Evaluation metrics}
Since our output space $\sa$ consists of three types of predictions and involves both discrete and continuous variables, we follow the metrics in~\cite{Wang_2019_CVPR,Liu_2020_CVPR}.
Specifically, as for binary variables $\saBin$ 
%(like the existence of sidewalk on the main road or crosswalk on the right side-road) 
and multi-class variables $\saMc$,
%(like the number of lanes on the right handside of ego car), 
the prediction accuracy is defined as
$\textrm{Accu.-Bi} = \frac{1}{14} \sum_{k=1}^{14} [p_k = \saBin{}_k]$ and $\textrm{Accu.-Mc} = \frac{1}{2} \sum_{k=1}^{2} [p_k = \saMc{}_k]$. We further report the F1 score on $\saBin$ to have a better idea about the overall performance given the observation that the binary classes are extremely biased. Formally, $\textrm{F1} = \frac{1}{14} \sum_{k=1}^{14} 2 \times \frac{p_k \times r_k}{p_k + r_k}$, where $p_k$ and $r_k$ are the precision and recall rate on $\saBin{}_k$ . For continuous variables, we report the mean square error (MSE). 
%Specifically, as for binary variables $\saBin$ (like the existence of sidewalk on the main road or crosswalk on the right side-road) and for multi-class variables $\saMc$ (like the number of lanes on the right handside of ego car), we report the prediction accuracy as $\textrm{Accu.-Bi} = \frac{1}{14} \sum_{k=1}^{14} [p_k = \saBin{}_k]$ and $\textrm{Accu.-Mc} = \frac{1}{2} \sum_{k=1}^{2} [p_k = \saMc{}_k]$. We further report the F1 score on $\saBin$ to have a better idea about the overall performance since we observe that the binary classes are extremely biased. Specifically, $\textrm{F1.-Bi} = \frac{1}{14} \sum_{k=1}^{14} 2 \times \frac{p_k \times r_k}{p_k + r_k}$, where $p_k$ and $r_k$ are the precision and recall rate on $k$-th variable of binary attributes. For regression variables we use the mean square error (MSE).  

Apart from parametric predictions, our model also outputs intermediate representations, e.g. OSP and HST. We further report the IoU as well as the accuracy for these two semantic segmentation tasks.~\buyu{Please note that human annotated OSP and HST are not available on either dataset in practice. Thus, we report our performance by comparing our predictions with rendered semantics $\bevmap$ and $x^p$ instead.}
% Please note that since we do not have human annotated per-pixel ground-truth for these two tasks, we report our performance by comparing our predictions with rendered semantics $\bevmap$ and $x^p$ instead.

% In addition, we also report Intersection-over-Union (IoU) as a overall performance evaluation metric. Specifically, we assume that we can render four-class semantic top-view maps with either the predicted results or the ground-truth annotations. Then we report the average IoU score over all test images. More details on IoU are presented in supplementary.

\subsection{Evaluations of Parametric Road Layout}
%on Single Image}
%Our experiments are conducted with a single image as input. 
\begin{figure*}[t]
 \setlength{\belowcaptionskip}{-0.35cm}
 \centering
  \includegraphics[width=1.0\linewidth]{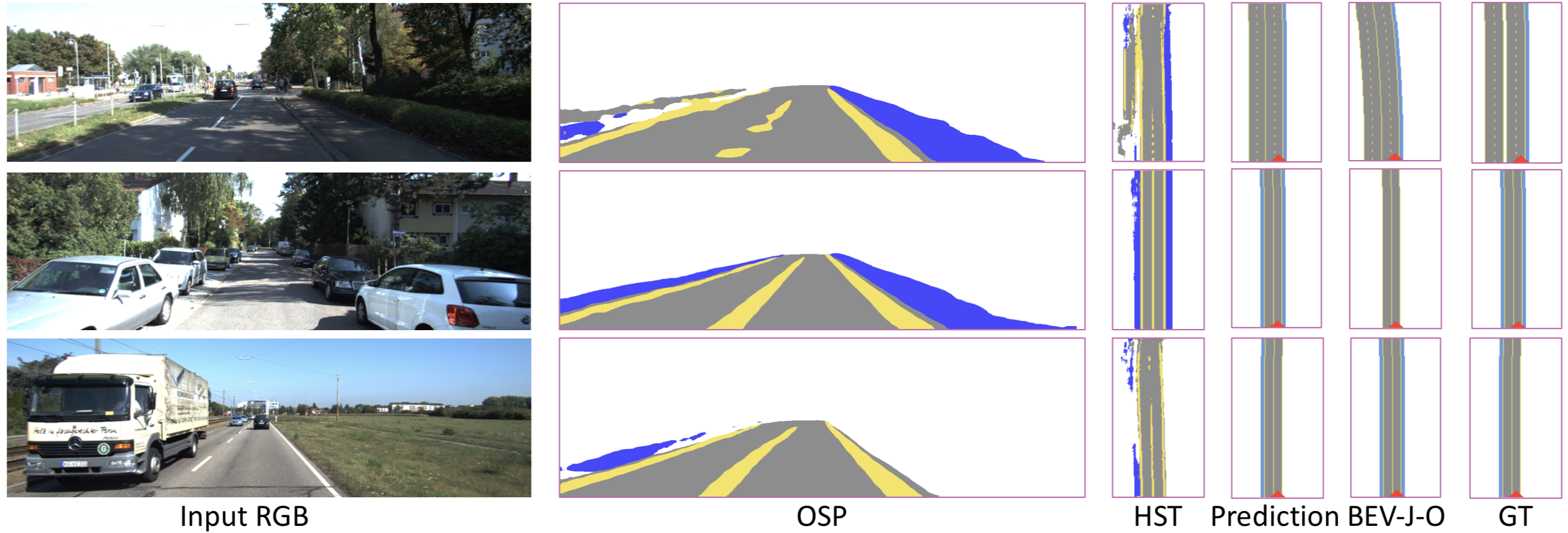}
  %\vspace{-0.2cm}
  \caption{Full predictions of our proposed model. From left to right: input RGB, OSP, HST, image rendered from parametric predictions, results from~\cite{Liu_2020_CVPR} and image rendered from ground-truth attributes. %\BB{for clarity, put "Input RGB" "OSP" "HST" "Prediction" "GT" on top of each column, so readers do not need to keep looking at caption and figure back and forth to figure out which is which.}
  }
  \label{fig:full_results}
\end{figure*}

\paragraph{Baselines}
We choose several appropriate baselines as presented in~\cite{Wang_2019_CVPR,Liu_2020_CVPR}.:
\begin{itemize}
\item \textbf{RGB} ($RGB$): A ResNet-101~\cite{sam:Seff16a,sam:He16a} backbone is introduces and trained on the manually-annotated ground truth. Note that this setup is the~\textbf{only} one that directly comparable to ours as it requires only the parametric annotations as ground-truth. %Seff and Xiao~\cite{sam:Seff16a} have the same setup except that we use a network with more parameters and train for all attributes simultaneously in a multi-task setup.
\item \textbf{RGB+Depth} ($RGB+D$): Same as $RGB$ but with the additional task of monocular pixel-wise depth prediction~\cite{sam:He16a}. 
%Note that in our proposal, 
In contrast, we do not require dense depth information. %Nor do we need any depth supervision.  
\item \textbf{BEV}($BEV$): $BEV$ uses the output of~\cite{sam:Schulter18a}, which is a top-view semantic map. To obtain such map, additional pixel-level semantic annotation and depth supervision are required in perspective space.~\buyu{Though more recent approaches~\cite{mani2020monolayout,philion2020lift} are also able to output semantics in top-view, they miss importance semantics such as lane boundary or crosswalk thus are not desired as $BEV$ baselines.}% while in our scenario, we requests neither.
\end{itemize}

We also report the performance of SOTA methods for single image top-view layout prediction, or~\textbf{H-BEV-DA}~\cite{Wang_2019_CVPR} and~\textbf{BEV-J-O}~\cite{Liu_2020_CVPR}. Please note that both of them require far more human annotations compared to our method.~\buyu{We refer the readers to supplementary for more details for all baselines.}
%request thousands of additional pixel-level semantic annotations and depth supervision while ours only requests cheap parametric annotations.

% \vspace{0.3cm}
% \noindent{\textbf{Quantitative results:~~}}
\paragraph{Quantitative results} Tab.~\ref{tbl:kitti_single_img} summarizes our main results on KITTI~\cite{sam:Geiger13a}. First of all, if we compare only to method with the same setting, or $RGB$, our method outperforms it with a large margin, which indicates the effectiveness of introducing PS and TS as intermediate modules. Furthermore, compared to $RGB+D$ method that introduces depth channel, or even the $BEV$ that further requires thousands of human labelled semantic segmentation images in perspective space, our method achieves better results, which is of significance given that our method requires far less human annotations. %supervision. %.% without time-consuming pixel-wise annotation, 
%which is actually very impressive since we do not require any pixel-wise annotation but is able to outperform methods that request more supervision. 
Note that both $H$-$BEV$-$DA$ and $BEV$-$J$-$O$ are based on $BEV$ but require even more human annotations. By comparing to $H$-$BEV$-$DA$ that further exploits additional simulated data and $BEV$-$J$-$O$ that requires the 3D object information as well as entire video sequence as input, we can see that our method achieves comparable results with far less
% \BB{The below timing shows key advantange of our annotations, why not bring those up earlier in the method section and should be in main text not footnote?}
human annotations.
%As reported in~\cite{Wang_2019_CVPRworkshop}, it takes about 20 minutes to annotate semantics of visible regions on KITTI perspective images, while the parameter annotations in BEV takes 20s per image on average. As for object cues, annotating single bounding box in 2D takes about 40s on average~\cite{Su_crowdsourcingannotations}, Annotating 3D bounding box as required in~\cite{Liu_2020_CVPR} can be more time consuming in practice.

We further report results on NuScenes~\cite{sam:NuScenes18a} in Tab.~\ref{tbl:nuscenes_updated}. Our method outperforms $RGB$ significantly. %we observe that our method is able to beat SOTA significantly when having the same model setting (See $RGB$ method). 
It also outperforms~\cite{Liu_2020_CVPR,Wang_2019_CVPR} with far less human annotations required.

\iffalse
\begin{table}
  \centering\small
  \input{tables/nuscenes_results}
  \caption{Results on NuScenes dataset. We observe that our method beats $RGB$ when having the same setting. Meanwhile, our results are better than SOTA ($H$-$BEV$+$DA$ and $BEV$-$J$-$O$) with far less human annotations required.}
  % with deep supervision protocol.}
  \label{tbl:nuscenes}
\end{table}
\fi

\begin{figure}
 \setlength{\belowcaptionskip}{-0.35cm}
 \centering
  \includegraphics[trim=420 0 0 0,clip,width=1.0\linewidth]{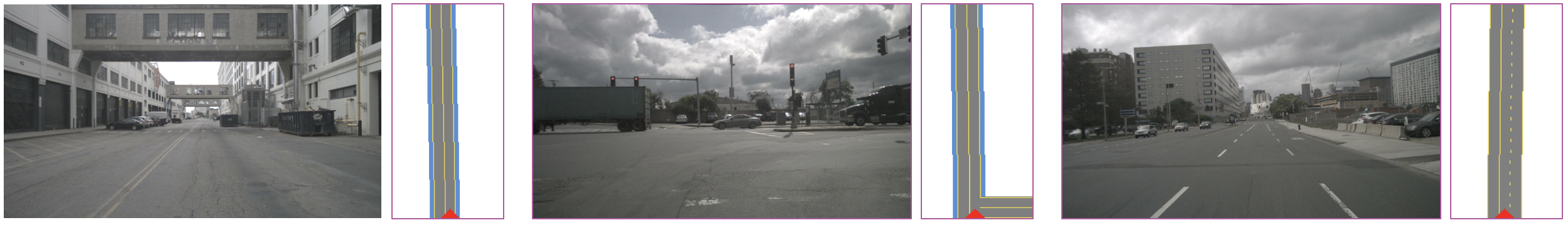}
  %\vspace{-0.5cm}
  \caption{Examples on NuScenes dataset. Left:input RGB Right: rendered BEV semantics from our prediction. %Our proposed method is able to handle diverse road layout, e.g. intersection or various number of lanes.
  }
  \label{fig:nuscenes_res}
\end{figure}

% \vspace{0.3cm}
% \noindent{\textbf{Qualitative results:~~}} 
\paragraph{Qualitative results} We demonstrate some qualitative results in Fig.~\ref{fig:full_results}. Note that in KITTI test sequences, rather than the road being occluded by cars driving in front, significant occlusions happen between parked cars and road/sidewalk, or between foreground classes, e.g. buildings or trees, and curved road or sideroad. As observed in this figure, our model is able to output satisfactory results on all three representations. We are able to handle complex road layout such as arbitrary number of lanes 
%,curved road and 
with heavy occlusions. Again, please note that OSP and HST %(the second and third column) 
are obtained without per-pixel human annotations. Our final layout prediction is also better than~\cite{Liu_2020_CVPR}. We further visualize our final results on NuScenes in Fig.~\ref{fig:nuscenes_res}. It shows that our model is able to handle various road layouts. We refer readers to supplementary material for more qualitative results.

%We refer readers to supplementary material for further studies in occlusion cases and more qualitative results on both KITTI and NuScenes dataset.
% \BB{Better to have qualitative results on NuScene. Just one example will do; it is 0 to 1 difference to the submission.}

% \vspace{0.3cm}
\vspace{0.3cm}

\begin{table*}\footnotesize
  
  \begin{minipage}{0.4\linewidth}  
  \setlength{\tabcolsep}{1.5pt}
  \begin{tabular}{l|cccc}
  %\hline
  %            & \multicolumn{4}{c}{NuScenes~\cite{sam:NuScenes18a}} \\
  \hline
  Method      & Accu.-Bi. $\uparrow$ &  Accu.-Mc. $\uparrow$ & MSE $\downarrow$ & F1 $\uparrow$ \\
  \hline
  RGB~\cite{sam:Seff16a,he2016deep} & .850 & .503 & .084 & .109\\
  \hline
  BEV~\cite{Wang_2019_CVPR} & .846 & .485 & .073 & .101\\
  H-BEV+DA~\cite{Wang_2019_CVPR}+GM & \textbf{.877} & .496 & .032 & .125\\
  \hline
  BEV-J-O~\cite{Liu_2020_CVPR}  & .858 & \underline{.543}  & \underline{.027} & \underline{.128} \\  
  \hline
  Ours & \underline{.875} & \textbf{.560} & \textbf{.023} & \textbf{.130} \\    
  \hline
\end{tabular}
  \caption{Results on NuScenes dataset. We observe that our method beats $RGB$ significantly when having the same model setting. Meanwhile, it also outperforms $H$-$BEV$+$DA$ and $BEV$-$J$-$O$ with far less human annotations required.}
  \label{tbl:nuscenes_updated}
  \end{minipage}
  \hspace{0.4cm}
  \begin{minipage}{0.55\linewidth}  
  \setlength{\tabcolsep}{1.5pt}
  \begin{tabular}{l|c|c|c|c|cccc}
  \hline
              &
              \multicolumn{4}{c|}{Module} &\multicolumn{4}{c}{KITTI~\cite{sam:Geiger13a}} \\
  \hline
  Method & $f^{ps}$ & $f^{trans}$ & $f^{halln}$ & $f^{tpp}$ &  Accu.-Bi. $\uparrow$ & Accu.-Mc. $\uparrow$ & MSE $\downarrow$ & F1 $\uparrow$ \\
  \hline
  RGB & & & & \checkmark& .811 & .778 & .230& .176\\
  \hline
  RGB+PS  & \checkmark & & & \checkmark & .822  & .827  & .159&  .425\\
  \hline
  RGB+PS+T & \checkmark & \checkmark &  & \checkmark & .826 & .829 & .144 & 441 \\ 
  \hline
  Ours & \checkmark & \checkmark & \checkmark & \checkmark& .833 & .832 & .140 & .473 \\
  \hline
\end{tabular}
  \caption{Ablation study on single image based road layout prediction on KITTI. Note that all these methods share the same amount of human annotations. We can see that our introduced PS and TS modules, on the one hand, provide meaningful intermediate representations at no additional costs. On the other hand, they also prove to be beneficial individually for the final parametric prediction task.}
  \label{tbl:kitti_ablation_updated}
  \end{minipage}
\end{table*}

\begin{table*}\small
  \centering
  \begin{tabular}{l|cc|c|c|c|c|c|c|c|c|c|c||c|c}
  \hline
  &\multicolumn{2}{c|}{Representation} 
  &\multicolumn{11}{c}{KITTI~\cite{sam:Geiger13a}} \\
  \hline
  \multicolumn{1}{c|}{\multirow{2}{*}{Data}} & \multicolumn{1}{c}{\multirow{2}{*}{OSP}}  & \multicolumn{1}{c|}{\multirow{2}{*}{HST}} & \multicolumn{2}{c|}{Road} & \multicolumn{2}{c|}{Land Boundary} & \multicolumn{2}{c|}{Sidewalk} &  \multicolumn{2}{c|}{Crosswalk} & \multicolumn{2}{c||}{Foreground} & \multicolumn{2}{c}{Average} \\
  \cline{4-15}
   & &  & Accu. & IoU & Accu. & IoU & Accu. & IoU & Accu. & IoU &  Accu. & IoU & Accu. & IoU \\
  \hline
  RGB+PS  & \checkmark & & .689 & .563 & .365 & .214 & .226 & .126 & .010 & .007 & .954 & .878 & .449 &.358 \\
  \hline
  Ours  & \checkmark & & .700 & .605 & .403 & .272 & .255 & .147 & .042 & .033 & .962 & .883 & .472 & .388 \\
  & & \checkmark & .605 & .461 & .272 & .197 & .167 & .102 & .038 & .032 & .868 &.651 &.390 & .289 \\
  \hline
\end{tabular}
  \caption{Intermediate results on KITTI. We report both IoU and accuracy for each semantic category. Compared to $RGB$+$PS$, our method achieves better performance in terms of $OSP$ with the help of end-to-end training. Our method further provides meaningful $HST$ results.}
  \label{tbl:intermediate_results_accuracy_iou}
\end{table*}

\subsection{Ablation Study}~\label{sec:ablation}
To demonstrate the effectiveness of intermediate modules as well as deep supervision, we further conduct experiments on incrementally adding modules. $RGB$ is the one without any module. $RGB$+$PS$ contains the PS module and directly predicts parametric predictions with perspective outputs. Formally, $RGB$+$PS$ is formulated as:
\begin{equation}
    \sa = \nnFull^{\textrm{rbgp}}(\im)=(\nnFull^{tpp} \circ \nnFull^{\textrm{ps}})(\im) \;,
\end{equation}
% where $\nnFull$ and $\nnFull^{\textrm{ps}}$ corresponds to perspective and parametric prediction modules in $RGB$+$PS$. % model $\nnFull^{\textrm{rbgp}}$.

%We further the 
Similarly, $RGB$+$PS$+$T$ is formulated as:
\begin{equation}
    \sa = \nnFull^{\textrm{rbgpf}}(\im)=(\nnFull^{tpp} \circ \nnFull^{\textrm{trans}} \circ \nnFull^{\textrm{ps}}   )(\im) \;,
\end{equation}
%where $\nnFull^{\textrm{trans}}$ corresponds to the feature transformation and $RGB$+$Persp.$+$Trans.$ model is denoted as $\nnFull^{\textrm{rbgpf}}$.

We report the quantitative results in Tab.~\ref{tbl:kitti_ablation_updated}. The results show that first of all, comparing the $RGB$ to $RGB$+$PS$, perspective representation, or the OSP, is beneficial for improving final parametric predictions. Secondly, the performance gap between $RGB$+$PS$+$T$ and $RGB$+$PS$ demonstrates the effectiveness of introducing top-view semantics as intermediate representation. Finally, by comparing the full model with $RGB$+$PS$+$T$, we can tell that the hallucination module is also critical for layout prediction task.
%we can tell that the hallucination module is also critical to further boost the final performance.
%we can tell that the hallucination module is beneficial in improving the top-view semantics, or HST, thus further boost the performance of our parametric layout prediction tasks.
%\BB{These are not ablation study. Move to Sec. 4.1.}

%\BB{Why put this in "Ablation study" section? Can give a new section "Annotation Time" or move to before Sec.4.1 with a heading "Annotation time"}

%\noindent{\textbf{Occlusion study:~~}} 
\paragraph{Occlusion study} Here, we study performance against increasing number of objects in the scene, indicating increasingly severe occlusions. Since~\cite{Wang_2019_CVPRworkshop,sam:Geiger13a} do not provide pixel-level semantic ground-truths on our test sequences, to analyze our ability to handle occlusions, we instead report the average image-level IoU on four classes against number of foreground objects with respect to rendered ground truth $x$ in Tab.~\ref{tbl:stat_obj}, with objects detected by Stereo-RCNN~\cite{li2019stereo}. As one can see, our method outperforms the state-of-the-art consistently, with increasing gap when having more objects.~\buyu{Please note that our model is single-image based so we handle all objects, no matter they are moving or not, in the same manner.}

\begin{table}
  \setlength{\tabcolsep}{4pt}
  \centering\small
  \begin{tabular}{l|ccccccccc|c}
  \hline
  Obj. & 0 & 1 & 2 & 3 & 4 & 5 & 6 & 7 & 8 & Avg. \\
  %\hline
  %&\multicolumn{10}{c}{Considering Four Semantic Classes} \\
  \hline
  ~\cite{Wang_2019_CVPR}  & .67 & .78 & .67 & .64 & .61 & .45 & .48 & .37 & .35 & .45\\
  Ours  & .78 & .81 & .72 & .69 & .67 & .48 & .50 & .40 & .35 & .50\\
  \hline
   &\multicolumn{10}{c}{Considering Road Class Only} \\
  \hline
  ~\cite{Wang_2019_CVPR}  & .85 & .74 & .85 & .79 & .67 & .63 & .59 & .57 & .37 & .63\\
  Ours  & .86 & .91 & .86 & .83 & .70 & .70 & .61 & .61 & .50 & .70\\
  \hline
\end{tabular}
  \caption{\small{Average per-image IoU w.r.t. number of road participants.}}
  % with deep supervision protocol.}
  \label{tbl:stat_obj}
\end{table}

\subsection{Evaluations of Intermediate Representations}~\label{sec:intermediate_results}
Apart from requiring less annotation while maintaining comparable performance, another advantage of the proposed method is being able to provide meaningful pixel-level intermediate representations, OSP and HST, as by-products. 
%without requiring per-pixel human annotations for either. 
To demonstrate that these intermediate representations are indeed semantically useful for downstream tasks, we  study their IoU as well as accuracy score, as an indication for their performance.
%on these intermediate representations. 
Please note that compared to existing work that requires dense and time-consuming pixel-wise human annotation, %ours can be treated as an unsupervised method 
ours only requires cheap parametric human annotations and produces pixel-level occlusion-reasoned semantic segmentation in perspective and top-view. 

As shown in Tab.~\ref{tbl:intermediate_results_accuracy_iou}, our method is able to provide multiple meaningful intermediate representations. Also, our deep supervision proves to be beneficial in an end-to-end manner, which can be observed from the performance gap on OSP between $RGB$+$PS$ and our full model. 
In addition, our model also achieves reasonably good performance on HST. As a reference, \cite{roddick2020predicting}, which  aims to predict pixel-level semantics of visible regions in top-view with perspective images as input, reports about $63.0\%$ IoU for drivable category on two different datasets.  

%pixel-level occlusion-reasoned semantic segmentation in perspective and top-view.
%without requiring per-pixel human annotations for either. 

\begin{figure}[t]
 \centering
  \includegraphics[width=1.0\linewidth]{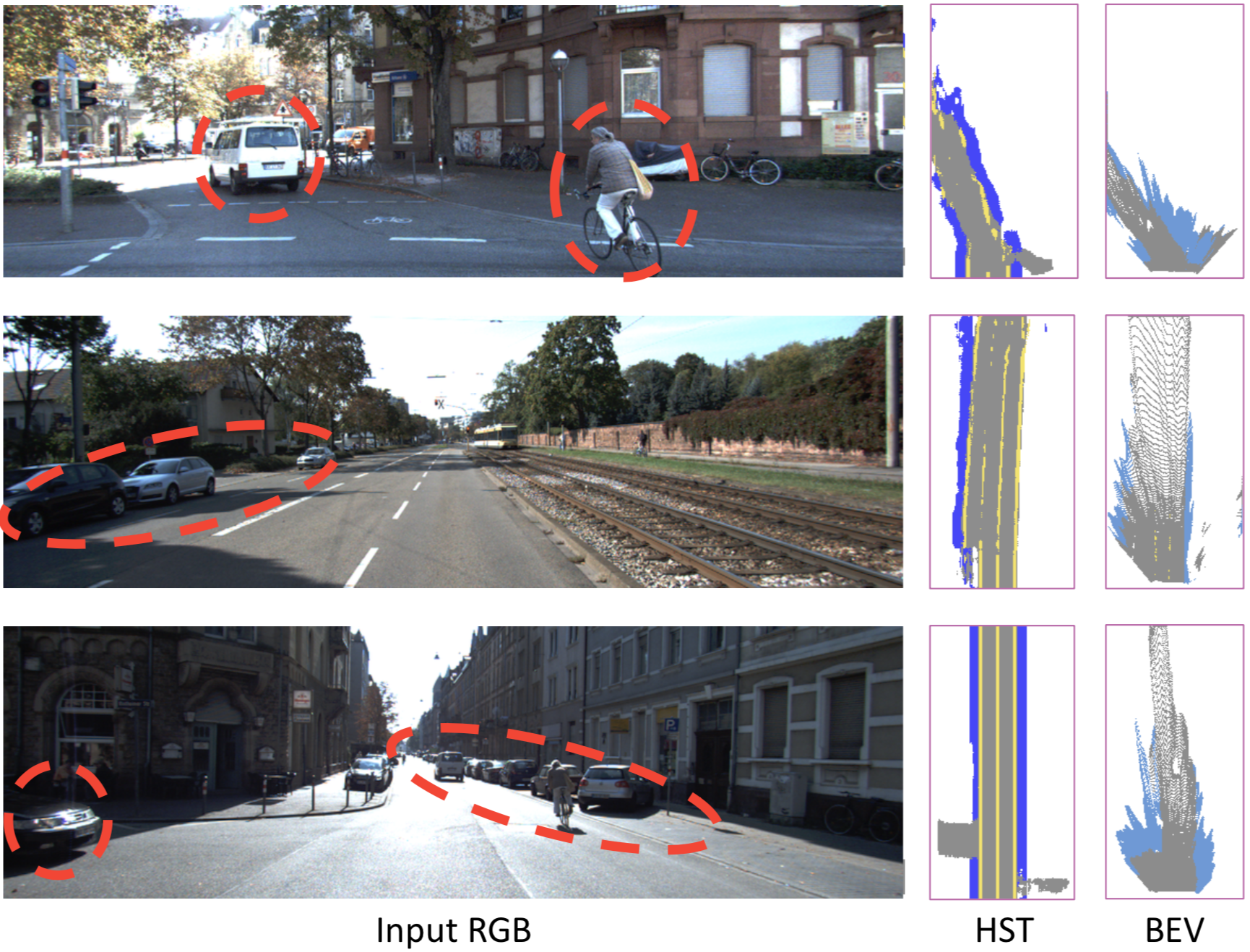}
  %\vspace{-0.5cm}
  \caption{We demonstrate input RGB, predicted HST as well as BEV of~\cite{Wang_2019_CVPR}, which is trained with thousands of pixel-level annotated images and LiDAR images. As can be seen in these examples, our model is able to hallucinate far away regions in a realistic manner, even on curved road, with $NO$ pixel-level human annotations. %Please note that all these results are obtained with $NO$ pixel-level human annotation. %\BB{put "Input RGB""Ours""[37]" on top of each column.}
  }
  \label{fig:semantics_bev}
\end{figure}

\begin{figure}[t]
 \setlength{\belowcaptionskip}{-0.35cm}
 \centering
  \includegraphics[width=1.0\linewidth]{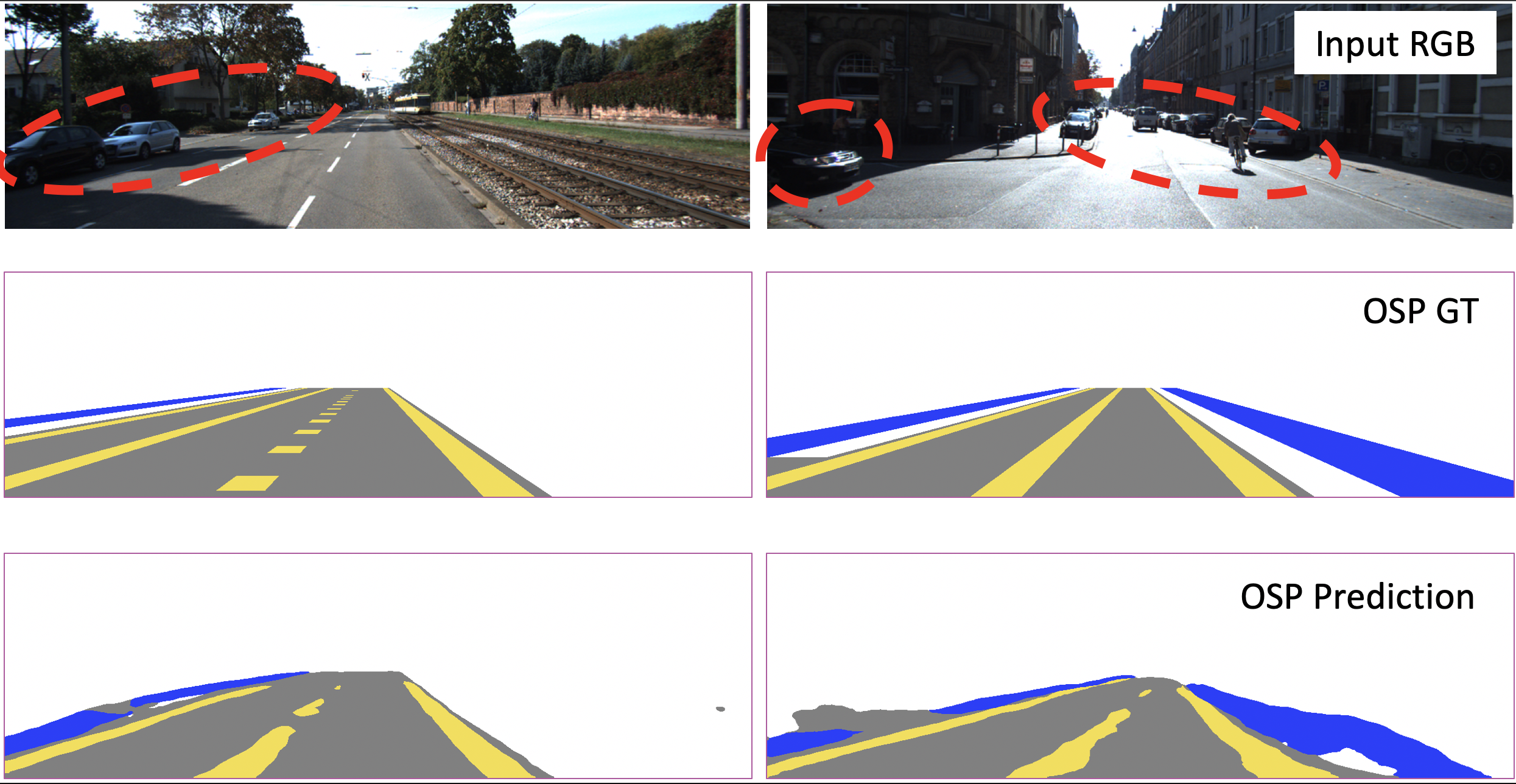}
  %\vspace{-0.5cm}
  \caption{Input image, generated ground-truth pixel-level semantics and predicted semantics from top to bottom row. Our model is able to predict the semantics quite well despite occlusions.
  %, even on regions that has been occluded by objects, e.g. cars, as highlighted. %\BB{Again, put "Input RGB""GT""Prediction" vertically on the left for each row. Also, highlight the occlusion regions using red circles.} %Please again note that all these results are obtained with $NO$ pixel-level human annotation. 
  }
  \label{fig:semantics_persp}
\end{figure}

%To demonstrate that these intermediate representations are indeed semantically useful for downstream tasks, we  study their IoU as well as accuracy score, as an indication for their performance.Please note that compared to existing work that requires dense and time-consuming pixel-wise human annotation, ours only requires cheap parametric human annotations and produces pixel-level occlusion-reasoned semantic segmentation in perspective and top-view. 
However, please note that~\cite{roddick2020predicting} requires pixel-level dense annotations in top-view during training and the predictions are not occlusion-reasoned. We further visualize quantitative results in Fig.~\ref{fig:semantics_bev} and Fig.~\ref{fig:semantics_persp}. As can be seen, our method obtain high quality semantics in both perspective and top-view despite occlusions.

%As shown in Tab.~\ref{tbl:intermediate_results_accuracy_iou}, the segmentation accuracy is reasonable given our setting. And we also observe that the end-to-end training improves the performance OSP as well by comparing our full method with $RGB$+$PS$. %We also report the IoU of these two methods in the same table. %The same observation can also be found in Tab.~\ref{tbl:intermediate_results_accuracy_iou} where we report the IoU metrics. 
%\footnote{\cite{roddick2020predicting} reports about $63.0\%$ IoU for drivable category in a 50m $\times$ 50m top-view space on two different datasets, at a resolution of 25cm per pixel. Ours is in 30m $\times$ 60m space represented by a 256 $\times$ 128 top-view map. Their results are obtained with pixel-level dense annotations in top-view and are not occlusion-reasoned.}. 
%Specifically,~\cite{roddick2020predicting} reports about $63.0\%$ IoU for drivable category on two different datasets.
%in a 50m $\times$ 50m top-view space on two different datasets, at a resolution of 25cm per pixel. %Ours is in 60m $\times$ 30m space represented by a 128 $\times$ 256 top-view map. Their results are obtained with pixel-level dense annotations in top-view and are not occlusion-reasoned
%We further visualize quantitative results of HST in Fig.~\ref{fig:semantics_bev}. As can be seen, our method obtain high quality semantics in top-view despite occlusions. 
% More qualitative results of OST can be found in supplementary materials.

\section{Conclusion}~\label{sec:conclusion}
We propose a novel end-to-end model that inputs single RGB perspective image and outputs multi-aspect representations for road layout, including top-view parametric predictions, OSP and HST. Specifically, we introduce two intermediate modules
%, PS and TS, 
and exploit deep supervision to learn inductive biases in occlusion-reasoning, geometric transformation and semantic abstraction. 
%To achieve that, we exploit deep supervision with very cheap parametric annotations only.  %without any per-pixel human supervision.%Our proposed method exploits the last two as meaningful intermediate representations without requesting labor-intense human annotations for either. 
We demonstrate the effectiveness of our proposed method as well as intermediate modules on publicly available datasets and demonstrate that we can achieve SOTA performance with less human annotations.
%beat the SOTA with a large margin when having the same setting and achieve comparable results w.r.t. SOTA while these methods require thousands of images to be densely annotated. 
% We also conduct detailed experiments to demonstrate the effectiveness and usefulness of intermediate modules.

%%%%%%%%% REFERENCES
{\small
\bibliographystyle{ieee_fullname}
\bibliography{egbib,myshortstring}
}

\end{document}